\documentclass[sigconf,authorversion,review=False, timestamp=False, nonacm, screen,pbalance]{acmart}

\AtBeginDocument{%
  \providecommand\BibTeX{{%
    \normalfont B\kern-0.5em{\scshape i\kern-0.25em b}\kern-0.8em\TeX}}}

\usepackage{tcolorbox}

\usepackage[maxfloats=30,morefloats=12]{morefloats}

\acmConference[WWW '24]{Proceedings of the ACM Web Conference 2024}{May 13--17,
  2024}{Singapore}
\acmPrice{15.00}
\acmISBN{978-1-4503-XXXX-X/18/06}

\begin{document}

\title{Improving Minority Stress Detection with Emotions}

\author{Jonathan Ivey}
\email{jwi001@uark.edu}
\affiliation{%
  \institution{University of Arkansas}
  \city{Fayetteville}
  \state{Arkansas}
  \country{USA}
  \postcode{72701-1201}
}

\author{Susan Gauch}
\email{sgauch@uark.edu}
\affiliation{%
  \institution{University of Arkansas}
  \city{Fayetteville}
  \state{Arkansas}
  \country{USA}
  \postcode{72701-1201}
}

\renewcommand{\shortauthors}{Ivey and Gauch}

\begin{abstract}
Psychological stress detection is an important task for mental healthcare research, but there has been little prior work investigating the effectiveness of psychological stress models on minority individuals, who are especially vulnerable to poor mental health outcomes. In this work, we use the related task of minority stress detection to evaluate the ability of psychological stress models to understand the language of sexual and gender minorities. We find that traditional psychological stress models underperform on minority stress detection, and we propose using emotion-infused models to reduce that performance disparity. We further demonstrate that multi-task psychological stress models outperform the current state-of-the-art for minority stress detection without directly training on minority stress data. We provide explanatory analysis showing that minority communities have different distributions of emotions than the general population and that emotion-infused models improve the performance of stress models on underrepresented groups because of their effectiveness in low-data environments, and we propose that integrating emotions may benefit underrepresented groups in other mental health detection tasks.
\end{abstract}





\maketitle

\section{Introduction}
Psychological stress detection from social media posts has been identified as an important task for mental healthcare research \citep{turcan-mckeown-2019-dreaddit}, but the datasets for this task may not fairly represent all groups, and little prior work has investigated the effectiveness of psychological stress models on minority individuals.

This issue is especially relevant for sexual and gender minority (SGM) people, who are more vulnerable to poor mental health outcomes than the general population. They are at higher risk of mental illnesses and suicide \citep{diaz_impact_2001, remafedi_relationship_1998, tebbe_suicide_2016}, and social media is often a place where SGM people find peers, seek help, and cope with prejudice \citep{formby_exploring_2017, pullen_safe_2014, mcinroy_its_2020, woznicki_parasocial_2021}.

\begin{figure}
    \begin{tcolorbox}[colback=white,colframe=black]
    At school, I have great friends and a good family at home. But I’m a closeted gay. If I ever came out, I know my friends would never talk to me again and my family would disown me. Because of this, I have zero motivation to come out.
    \end{tcolorbox}
    \begin{tcolorbox}[colback=white,colframe=black]
    I was kicked out of my online video game squad just for being gay (they said gays are pedophiles). My dad was outside my room listening when this happened so he grabbed me by the throat and kicked me out.
    \end{tcolorbox}
    \caption{Examples of minority stress disclosure on social media from \citet{saha_language_2019}.}
    \label{fig:minority-stress-example}
\end{figure}
 
One way to evaluate the ability of psychological stress models to understand the language of SGM individuals is through the related task of minority stress detection. Like psychological stress detection, minority stress detection uses natural language processing techniques to classify social media posts with whether the poster is experiencing stress \citep{saha_language_2019, cascalheira_classifying_2022}. However, minority stress is a psychosocial stress specific to minority individuals that they experience due to stigmatized social status \citep{meyer_prejudice_2003}. An example of minority stress on social media is provided in Figure \ref{fig:minority-stress-example}.

This task has an important application in improving the methodology of minority stress studies by circumventing limitations in survey-based self-reporting \citep{heppner_research_2016}. The systematic detection of minority stress can also be used to study large-scale health trends on social media that are not feasible to collect survey data on. Additionally, it has applications in automated intervention for those at risk of adverse consequences and screening for comorbid risks like cancer, HIV, and reduced cardiovascular health \citep{flentje_relationship_2020}.

In this work, we evaluate the effectiveness of psychological stress models at detecting minority stress, and we hypothesize that a lack of diversity in the psychological stress training data causes stress models to overfit and be unable to generalize to minority individuals.

To address this issue, we experiment with the multi-task emotion-infused architectures introduced by \citet{turcan-etal-2021-emotion}. They explored connections between emotions and stress in deep learning models, and they demonstrated that the task of emotion detection, which has more labeled data available, could improve the explainability of stress models.

At their introduction, emotion-infused models did not improve performance on the psychological stress detection task; however, we note that multi-task learning techniques (like those used for the emotion-infused models) are also known to improve generalization \citep{caruana_multitask_1997}. In this work, we explore using multi-task emotion-infused models to improve minority stress detection and highlight their potential for improving the performance of other mental health models on minorities. Our contributions in this work are as follows: 
\begin{itemize}
\item We conduct experiments to demonstrate that traditional single-task psychological stress detection models underperform on minority stress and highlight how this performance difference risks widening preexisting healthcare disparities experienced by minority communities. 

\item We demonstrate that emotion-infused models reduce the performance gap and exceed state-of-the-art performance for the minority stress detection task without training on minority stress data.

\item We provide explanatory analysis showing that minority communities have different distributions of emotions than the general population and that emotion-infused models improve the performance of stress models on underrepresented groups because of their effectiveness in low-data environments.
\end{itemize}

\section{Related Work}
\subsection{Psychological Stress Detection}
Psychological stress detection is best studied with physiological data. Prior work has used audio \citep{zuo-etal-2012-multilingual}, biological markers \citep{allen_biological_2014}, neuroimaging \citep{al-shargie_mental_2016}, thermal imaging \citep{kumar_stressnet_2020}, or combinations of these signals \citep{jaiswal-etal-2020-muse} to achieve the most accurate forms of psychological stress detection. However, \citet{turcan-mckeown-2019-dreaddit} demonstrated the value and feasibility of detecting psychological stress purely from social media text. 

\citet{turcan-etal-2021-emotion} later published a work that introduced the use of emotion-infused models for psychological stress detection. These models improved explainability by integrating emotion with multi-task learning or fine-tuning; however, they did not significantly improve the performance of psychological stress detection.

The authors also acknowledged demographic imbalances in the psychological stress dataset and noted a lack of language representing minority groups; however, little previous work has explored the performance of psychological stress models on minorities. In this work, we use the minority stress detection task to highlight the limitations of traditional architectures for detecting stress in minority individuals and explore the benefits of the previously introduced emotion-infused architectures for overcoming those limitations.

\subsection{Minority Stress Detection}
\citet{saha_language_2019} is the first work to introduce the use of natural language processing techniques for understanding minority stress. They wrote a codebook for identifying minority stress, created the first dataset of social media posts annotated for disclosure of minority stress, and introduced the first machine learning classifier for minority stress on social media. They experimented with using expertly engineered language features in combination with machine learning models to build a classifier. Their models are the current state-of-the-art for minority stress detection.

Building off that work, \citet{cascalheira_classifying_2022} introduced a proof-of-concept BI-LSTM to detect minority stress without expertly engineered features. They were the first to use deep learning for this task; however, they found limited results that did not outperform traditional machine learning models in detecting minority stress.

\citet{cascalheira_classifying_2022} viewed the minority stress detection task in isolation and trained models directly on the minority stress dataset (which is too small for deep learning models). In this work, we understand minority stress as a subset of psychological stress, and we use this framework to improve minority stress detection by improving models from the related task of psychological stress detection. 

\subsection{Pretrained Language Models}
Prior work has suggested that domain-specific pretrained language models may benefit mental healthcare tasks. \citet{ji-etal-2022-mentalbert} introduced the pretrained language models MentalBERT and MentalRoBERTa, which were trained on a corpus of 13,671,785 sentences from mental health-related subreddits. They demonstrated improved performance for psychological stress detection. However, little previous work uses pretrained language models for minority stress detection. In this work, we experiment with four pretrained language models (BERT, RoBERTa, MentalBERT, and MentalRoBERTa) paired with psychological stress models for minority stress detection. 

\section{Approach}
\subsection{Baselines}
For our baseline stress model, we use the pretrained BERT language model introduced \citet{devlin-etal-2019-bert} followed by an additional dropout layer and dense classification layer. This architecture is the simplest that we evaluate, and it performed the best on the psychological stress detection task when it was introduced in \citet{turcan-mckeown-2019-dreaddit}. Later architectures have not provided statistically significant performance improvements. We will refer to this model as Single-Task.

We also compare our minority stress models to the current state-of-the-art for minority stress detection established in \citet{saha_language_2019}. This model is a Multilayer Perceptron (MLP) algorithm trained with GloVe word embeddings \citep{pennington-etal-2014-glove}, LIWC psycholinguistic categories \citep{tausczik_psychological_2010}, a gender and orientation hate speech lexicon, n-grams, sentiment classification \citep{manning-etal-2014-stanford}, and stress classification \citep{saha_modeling_2017}. This series of expertly engineered features draws information from a wide range of data sources that consider lexical and semantic aspects of the text, with a special focus on LGBTQ+ issues.

\subsection{Emotion-Infused Models}
We hypothesize that integrating emotions into psychological stress models will account for differences in the emotional expressions of minority individuals and improve the models’ ability to generalize to minority stress. To test this claim, we evaluate the emotion-infused models introduced in \citet{turcan-etal-2021-emotion}. When previously evaluated on psychological stress detection, these models did not provide significant performance improvements, but they improved explainability, and they represent key methods for using emotion in mental health tasks. There are three emotion-infused models.

\subsubsection{Fine-Tuning Model}
\begin{figure}[ht]
  \centering
  \includegraphics[width=\linewidth]{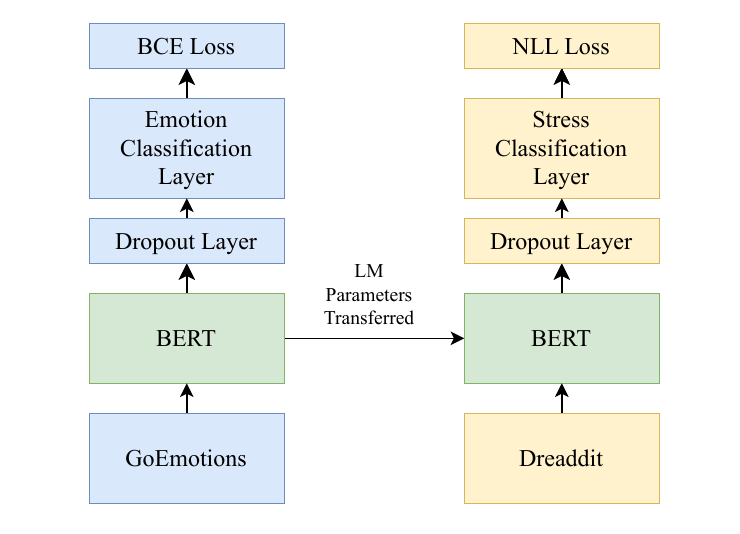}
  \caption{The architecture of Fine-Tune. Components that are used for both tasks are highlighted in green.}
  \label{fig:fine-tune}
\end{figure}
The first emotion-infused model that we evaluate is Fine-Tune. A visualization of Fine-Tune is shown in Figure \ref{fig:fine-tune}. In this architecture, we first fine-tune a single-task model for emotion detection. Because that is a multi-label task, the model trains using binary cross-entropy (BCE) loss. Then the language model parameters from that BERT model are transferred to another single-task model that is further fine-tuned for psychological stress detection. Because that is a single-label task, the model trains using negative log-likelihood (NLL) loss. The rationale for this architecture is that the first task would give the BERT language model a better understanding of emotions, and that understanding would enable a more holistic representation of stress.

\subsubsection{Alternating Multi-Task Model}
\begin{figure}[ht]
  \centering
  \includegraphics[width=\linewidth]{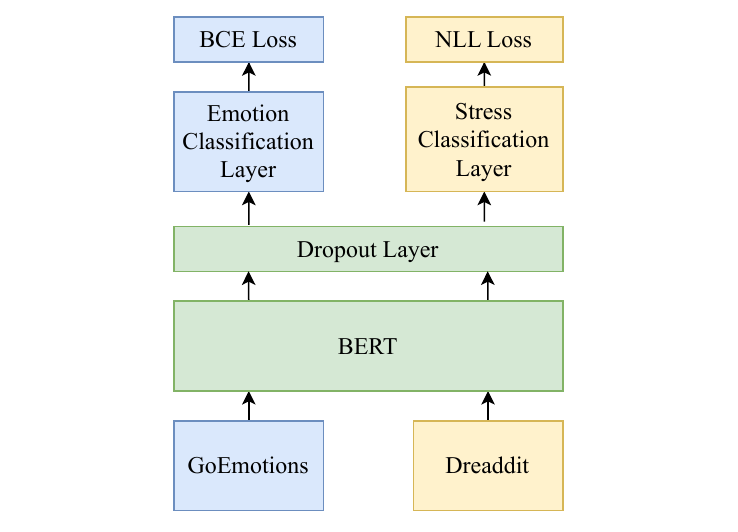}
  \caption{The architecture of Multi\textsuperscript{\textit{Alt}}. Components that are used for both tasks are highlighted in green.}
  \label{fig:multi-alt}
\end{figure}
The second emotion-infused model that we evaluate is Multi\textsuperscript{\textit{Alt}}. A visualization of Multi\textsuperscript{\textit{Alt}} is shown in Figure \ref{fig:multi-alt}. It follows a similar rationale as Fine-Tune, but instead of training and then transferring a separate language model, it trains a single, shared language model. During training, it alternates between training for emotion detection and psychological stress detection. Each training batch switches which task it is training for, but these different tasks share the same BERT representation layer. As in the Fine-Tune model, the emotion model trains with BCE loss and the psychological stress model trains with NLL loss.

\subsubsection{Classical Multi-Task Model}
\begin{figure}[ht]
  \centering
  \includegraphics[width=\linewidth]{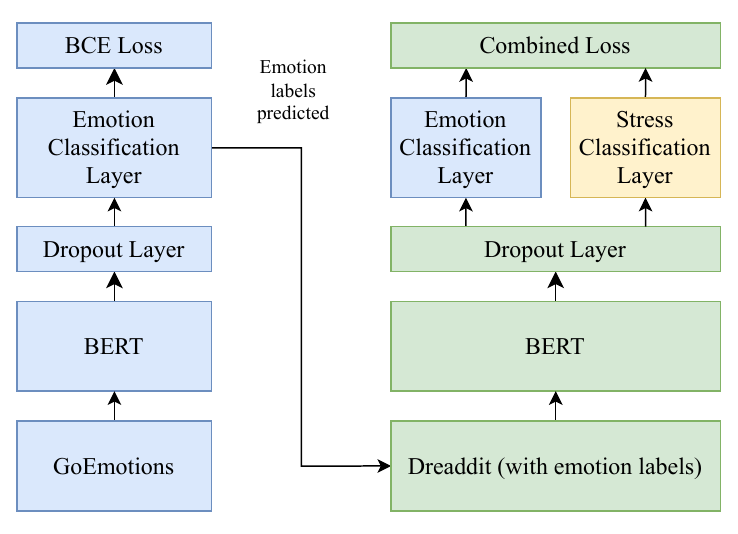}
  \caption{The architecture of Multi. Components that are used for both tasks are highlighted in green.}
  \label{fig:multi}
\end{figure}
The final emotion-infused model that we evaluate is Multi. A visualization of Multi is shown in Figure \ref{fig:multi}. It differs from the other two models because it uses a classical multi-task architecture that has the same input data for both tasks. However, because the stress data is not labeled with emotions, we first separately train a single-task model for emotion detection and use it to predict emotion labels for the stress data. The multi-task model then uses these emotion labels as targets for training the emotion detection task. In this model the loss is given by $\mathfrak{L}=\lambda\mathfrak{L}_{stress}+(1-\lambda)\mathfrak{L}_{emotion}$ where $\mathfrak{L}_{stress}$  is the NLL loss for psychological stress detection, $\mathfrak{L}_{emotion}$ is the BCE loss for emotion detection, and $\lambda$ is a weight parameter that we tune during model selection. The rationale for this model follows the traditional understanding that the inductive bias from the emotion detection task would improve its generalization.

\subsection{Pretrained Language Models}
Single-Task, Fine-Tune, Multi\textsuperscript{\textit{Alt}}, and Multi were all originally introduced using BERT, but \citet{ji-etal-2022-mentalbert} demonstrated that their domain-specific pretrained language models, MentalBERT and MentalRoBERTa, improved performance on the psychological stress detection task. In this work we evaluate four pretrained models tested in their work (BERT, RoBERTa, MentalBERT, and MentalRoBERTa) paired with each of the previously mentioned stress models.

\section{Data}
\begin{table*}[t]
\centering
\caption{Results of the models evaluated on minority stress detection with different pretrained language models. The best result under each metric is bolded.}
\begin{tabular}{@{}l|cc|cc|cc|cc@{}}
\multicolumn{1}{c|}{} & \multicolumn{2}{c|}{\textbf{BERT}}       & \multicolumn{2}{c|}{\textbf{RoBERTa}} & \multicolumn{2}{c|}{\textbf{MentalBERT}} & \multicolumn{2}{c}{\textbf{MentalRoBERTa}} \\
\textbf{Model}                 & \textbf{F1}             & \textbf{Accuracy}       & \textbf{F1}            & \textbf{Accuracy}     & \textbf{F1}             & \textbf{Accuracy}       & \textbf{F1}              & \textbf{Accuracy}        \\
\hline
Single-task                          & 69.85 & 70.88 & 74.88  & \textbf{75.09} & 73.49 & 73.56 & 73.33 & 73.95 \\
Fine-Tune                            & 69.47 & 70.31 & 70.87  & 71.45 & 68.71 & 70.50 & 72.24 & 72.61 \\
Multi\textsuperscript{\textit{Alt}}  & 70.95 & \textbf{71.45} & 70.60  & 70.69 & 73.58 & \textbf{74.52} & 71.88 & 72.41 \\
Multi                                & \textbf{75.55} & 68.58 & \textbf{75.16}  & 69.35 & \textbf{75.58} & 72.99 & \textbf{78.53} & \textbf{74.52 }\\

\end{tabular}
\label{tab:result1}
\end{table*}

\begin{table*}[t]
\caption{Results of the models evaluated on psychological stress detection with different pretrained language models. The best result under each metric is bolded.}
\centering
\begin{tabular}{@{}l|cc|cc|cc|cc@{}}
\multicolumn{1}{c|}{} & \multicolumn{2}{c|}{\textbf{BERT}}       & \multicolumn{2}{c|}{\textbf{RoBERTa}} & \multicolumn{2}{c|}{\textbf{MentalBERT}} & \multicolumn{2}{c}{\textbf{MentalRoBERTa}} \\
\textbf{Model}                 & \textbf{F1}             & \textbf{Accuracy}       & \textbf{F1}            & \textbf{Accuracy}     & \textbf{F1}             & \textbf{Accuracy}       & \textbf{F1}              & \textbf{Accuracy}        \\
\hline
Single-task                          & 77.70 & 77.95 & 78.18 & 78.37 & 76.03 & 76.88 & 79.35 & 79.39 \\
Fine-Tune                            & 75.85 & 75.67 & 77.48 & 77.53 & 77.30 & 77.43 & 76.37 & 76.78 \\
Multi\textsuperscript{\textit{Alt}}  & \textbf{78.76} & \textbf{78.97} & 77.00 & 77.25 & \textbf{80.80} & \textbf{80.89} & 79.42 & 79.59 \\
Multi                                & 77.53 & 77.36 & \textbf{79.00} & \textbf{79.16} & 77.90 & 78.27 & \textbf{79.86} & \textbf{79.91} \\

\end{tabular}
\label{tab:result2}
\end{table*}
For psychological stress detection, we use Dreaddit \citep{turcan-mckeown-2019-dreaddit}, a dataset of 3,553 segments of Reddit posts from communities where stress is commonly disclosed. Each segment was labeled with whether the poster is expressing stress using crowdsourced annotation, requiring a majority vote from five annotators. We use this dataset to train the stress models and evaluate their performance on psychological stress detection.

To evaluate how the models generalize to minority stress detection, we use an existing dataset of 350 Reddit posts collected from LGBTQ+ communities by \citet{saha_language_2019}. These posts were manually labeled by the authors with whether they contain the disclosure of minority stress using a codebook built based on Meyer’s Minority Stress model \citep{meyer_prejudice_2003}. In this paper, we will refer to this dataset as MStress.

Finally, to train the emotion-infused models, we use both Dreaddit and the GoEmotions dataset \citep{demszky-etal-2020-goemotions}. GoEmotions consists of 58,009 Reddit comments labeled by crowd workers with one of more than 27 emotions (or neutral). Based on the findings of \citet{turcan-etal-2021-emotion} we use a relabeling of this dataset created with agglomerative clustering to cluster the original labels into the Ekman 6 basic emotions (anger, disgust, fear, joy, sadness, surprise, neutral) \citep{ekman_are_1992}.

\section{Experimental Setup} \label{sec:experimental-setup}

Due to the scarcity of minority stress data, we do not train our models directly on MStress. Instead, all the models are trained on Dreaddit and GoEmotions (for the emotion-infused models) with minibatch gradient descent using the Adam optimizer \citep{kingma2017adam}.

We test on Dreaddit to evaluate their psychological stress detection, and we test on MStress to evaluate their minority stress detection. For Dreaddit and GoEmotions, we use 60\% of the data for training, 20\% for hyperparameter tuning, and 20\% for testing.

We replicate \citet{turcan-etal-2021-emotion}'s training and hyperparameter tuning processes with the same parameter ranges; however, for our primary tests, we use 50\% of MStress for hyperparameter tuning instead of using Dreaddit. We make this change to find the peak performance of the models for minority stress detection and highlight the associated reductions in psychological stress detection.

We run each of these experiments three times with three different random seeds, and we report the mean of the three runs. Further details about the training and hyperparameter tuning processes are available in Appendix \ref{sec:appendix} for reproducibility.

\section{Results}

We report the results of our primary tests when evaluated on minority stress in Table~\ref{tab:result1}, and we report the results of our primary tests when evaluated on psychological stress in Table~\ref{tab:result2}.

\subsection{Single-Task Models}
We find that the traditional models underperform on minority stress detection. The Single-Task models achieve F1 scores between 2.54 and 7.85 points lower on minority stress detection than on psychological stress detection. The MentalRoBERTa model performs best on psychological stress detection out of the Single-Task models with an F1 of 79.35, but that score drops to 73.33 when evaluated on minority stress detection.

\subsection{Emotion-Infused Models}
\begin{figure*}[t]
    \centering
    \includegraphics[width=\textwidth]{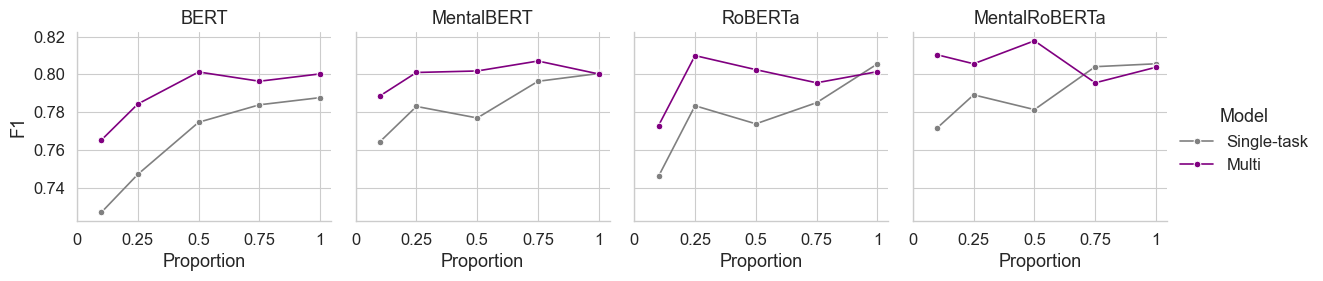}
    \caption{Performance of the Single Task and Multi models trained with different proportions (10\%, 25\%, 50\%, 75\%, and 100\%) of the original training set and evaluated on the psychological stress data.}
    \label{fig:data-reduction}
\end{figure*}
Though Fine-Tune and Multi\textsuperscript{\textit{Alt}} do not provide significant performance improvements, we find that the Multi models improve minority stress detection performance over baselines in all cases. Multi combined with MentalRoBERTa achieves an F1 of 78.53 on minority stress detection. While this result is still lower than the best psychological stress models, it closes the gap significantly and outperforms 13 out of the 16 psychological stress models. This result demonstrates that using the Multi architecture improves the ability of psychological stress models to generalize to minority stress detection. 

Importantly, our best Multi models are not trained directly on minority stress data, but they outperform the current state-of-the-art MLP for minority stress detection. The MLP proposed by \citep{saha_language_2019}, which is trained directly on minority stress data, achieved an F1 of 75, and our best Multi model achieves an F1 score 78.53.

\subsection{Pretrained Language Models}
We find that in most cases, domain-specific language models like MentalBERT and MentalRoBERTa provide marginal improvements in both psychological stress detection and minority stress detection compared to the standard BERT and RoBERTa models. This result confirms prior work demonstrating that the MentalBERT and MentalRoBERTa language models perform better on a variety of mental healthcare tasks \citep{ji-etal-2022-mentalbert}.

\subsection{Discussion}
These findings have important implications for the use of stress models in research and healthcare applications. First, traditional Single-Task models perform worse on minority stress detection than psychological stress detection and risk reinforcing preexisting mental healthcare disparities for SGM individuals. Second, the Multi architecture creates models that can generalize well to minority stress detection and significantly reduce the performance gap. Finally, minority stress researchers can benefit from using psychological stress detection data to surpass the current state-of-the-art without directly training on minority stress data.

In the next section, we explore justifications for these performance differences with an analysis of the Single-Task and Multi models for psychological stress detection with reduced training sets.

\section{Analysis}
\subsection{Data Reduction Analysis}
We propose that the disparity in performance of the baseline models between minority stress and psychological stress is due to overfitting on the psychological stress data. The Single-Task models gain too much sample-specific information and, as a result, are struggling to perform well on out-of-sample stress disclosures like minority stress.

Multi-task learning techniques improve generalization by using domain information contained in related tasks as an inductive bias \citep{caruana_multitask_1997}. We hypothesize that this improvement in generalization explains why the Multi models have improved performance on minority stress detection compared to baselines.

To further evidence this reasoning, we experiment with reducing the size of the Dreaddit training set for the psychological stress detection task. This reduction in the training set simulates the data scarcity that is present for minority stress detection.

We perform the same experimental setup as described in Section \ref{sec:experimental-setup}, but we use psychological stress data for our hyperparameter tuning and change the size of the training set to be either 10\%, 25\%, 50\%, 75\%, or 100\% of the original training set. We perform these experiments with the Single-Task and Multi models paired with each of the four language models (BERT, RoBERTa, MentalBERT, and MentalRoBERTa).

We report our results in Figure \ref{fig:data-reduction}. We find that while Single-Task and Multi achieve equivalent performance with the full training set, reduced training sets reduce the performance of Single-Task models much more significantly than Multi models. We see that at 100\% the Single-Task models all have F1 scores near 80, but at 50\% they drop to be between 77.38 and 78.13. By comparison, the Multi models have F1 scores of at least 80 with only 50\% of the training data, but they do not significantly improve as the training size increases.

This finding demonstrates that the improved generalizability of the multi-task architecture of the models makes them more effective in low-data environments. It consequently explains why Multi models are more effective at minority stress detection: the training set has a limited amount of minority stress, so detecting it is a low-data environment.

This finding also suggests that multi-task emotion-infused architectures may improve stress detection for other underrepresented groups, and further work should be done to explore using emotions to create equitable mental health models.

\subsection{Emotion Distributions}
\begin{figure*}[t]
    \centering
    \includegraphics[width=\textwidth]{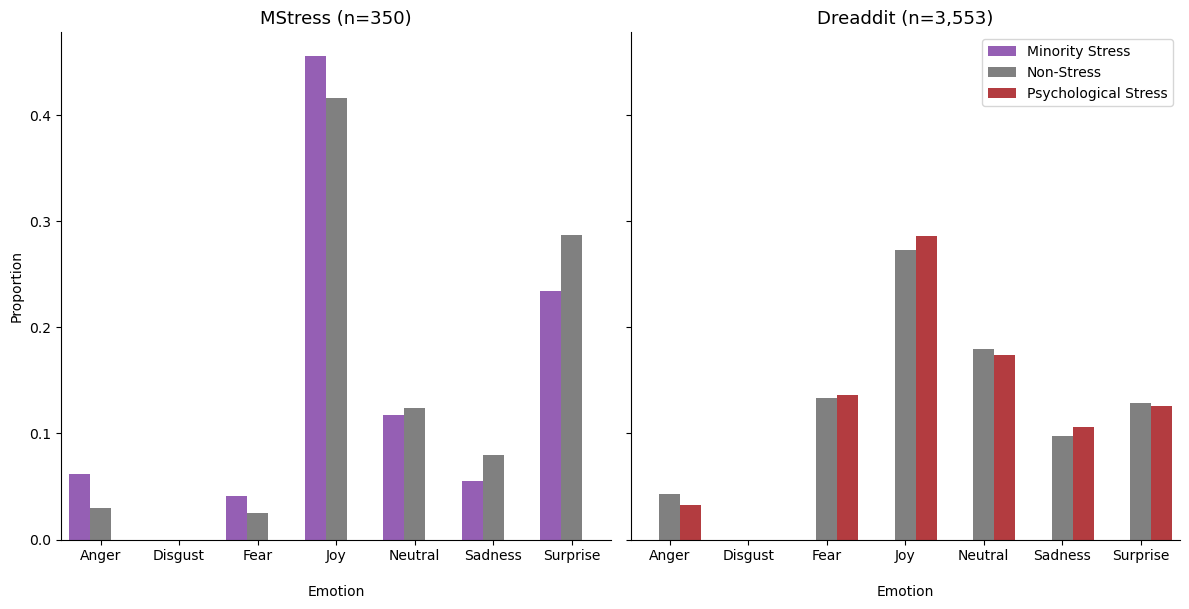}
    \caption{Distributions of predicted emotion labels in MStress and Dreaddit. Note that posts can have multiple emotion labels.}
    \label{fig:distributions}
\end{figure*}
To provide additional support for the importance of emotion analysis for supporting underrepresented groups, we examine the predicted emotion distributions of both MStress and Dreaddit (shown in Figure \ref{fig:distributions}). These emotion labels were created using a single-task MentalRoBERTa model with a macro F1 of 61.13.

From these distributions, we see that emotions do not significantly vary based on stress status. Posts in MStress marked as minority stress have a similar emotion distribution as posts marked as non-stress, and posts in Dreaddit marked as psychological stress have a similar emotion distribution as posts marked as non-stress.

However, we see a significant difference in the emotion distributions of posts from MStress compared to Dreaddit. This suggests that posts from minority communities exhibit different emotions, and that difference may affect single-task models’ ability to understand mental health conditions in minorities. \citet{Singh2023-ve}, which found that minority stress mediates emotion regulation, leading to dysregulation and emotion suppression, supports this finding.

This difference provides further explanation for the underperformance of single-task stress models on minorities. They are trained on a distribution of emotional expressions that are not representative of minority communities. This finding further suggests that multi-task, emotion-infused architectures may make more equitable models for other mental health tasks.

\section{Conclusion}

In this work, we find that traditional single-task stress models underperform on minority stress detection and are at risk of widening the healthcare gap for SGM individuals. We also find that risk can be reduced with the use of a multi-task architecture that integrates the task of emotion detection. Our experiments show that architecture performs well on psychological stress detection and outperforms the state-of-the-art for minority stress detection without training on minority stress data. Finally, we provide explanatory analysis demonstrating the Multi model’s superior performance in low-data environments, and we highlight how differences in emotion expressions in minority communities make them vulnerable to reduced effectiveness in mental health modeling. Our analysis suggests that integrating emotions may be effective for improving the performance of mental health models on underrepresented groups, and future work should explore using emotions to create equitable models for other mental health tasks.

\section{Limitations}
Our data was collected from Reddit, which is not necessarily representative of the general population. While our work focuses on generalizing stress models to work effectively on minority stress, it does not evaluate their ability to generalize to other social media platforms or beyond social media.

Additionally, our models focus on detecting minority stress for sexual and gender minorities, but we did not explore the detection of minority stress for other underrepresented groups like racial and ethnic minorities. Other minority groups may disclose minority stress differently, and future work should evaluate the performance of current minority stress models on other minority groups.

Finally, we note that because we chose to understand minority stress as a subset of psychological stress, our models are not able to differentiate between the two. Our framework requires that the models be combined with another method for identifying minority individuals before they can provide specific insights into the nature of minority stress.

\section{Ethical Considerations}

There are inherent risks involved with identifying minority stress. Bad actors who seek to target minority individuals would benefit from improved methods for detecting minority stress; however, we minimize that risk by building models that are effective at detecting both minority stress and psychological stress. That choice makes them ineffective for identifying minority individuals, so bad actors would not be able to use our models for that purpose.

Additionally, there is a risk that limitations in stress models could reinforce pre-existing healthcare disparities. Our work seeks to reduce existing disparities by expanding the capabilities of stress models, but as mentioned previously, our work is not comprehensive for all minority groups. Further evaluation must be done before these models can be used in real-world applications.

Finally, \citet{cascalheira_classifying_2022} identified the importance of consent when modeling stress in real-world applications. Many social media users benefit from anonymity, and they may not want their social media experience to change based on the stress classifications of their posts. This point is particularly relevant for SGM individuals who may not have come out yet and may not want their minority stress identified.


\bibliographystyle{ACM-Reference-Format}
\bibliography{custom, anthology}

\pagebreak
\appendix

\section{Appendix} \label{sec:appendix}
\begin{table*}[t]
\caption{Results of all the models on the MStress development set. The best result under each metric is bolded.}
\centering
\begin{tabular}{@{}l|cc|cc|cc|cc@{}}
\multicolumn{1}{c|}{} & \multicolumn{2}{c|}{\textbf{BERT}}       & \multicolumn{2}{c|}{\textbf{RoBERTa}} & \multicolumn{2}{c|}{\textbf{MentalBERT}} & \multicolumn{2}{c}{\textbf{MentalRoBERTa}} \\
\textbf{Model}                 & \textbf{F1}             & \textbf{Accuracy}       & \textbf{F1}            & \textbf{Accuracy}     & \textbf{F1}             & \textbf{Accuracy}       & \textbf{F1}              & \textbf{Accuracy}        \\
\hline
Single-task                          & 68.42 & \textbf{69.56} & 68.46 & 68.76 & 70.00 & 70.49 & 67.90 & 68.44 \\
Fine-Tune                            & 66.41 & 66.62 & 68.71 & 68.89 & 70.82 & 71.33 & 69.92 & \textbf{70.25} \\
Multi\textsuperscript{\textit{Alt}}  & 68.57 & 69.34 & 69.16 & 69.50 & 71.00 & \textbf{71.59} & 69.66 & 70.13 \\
Multi                                & \textbf{73.22} & 69.38 & \textbf{72.44} & \textbf{69.66} & \textbf{71.29} & 69.46 & \textbf{72.01} & 68.82 \\

\end{tabular}
\label{tab:dev-result}
\end{table*}
We provide the following information for reproducibility.

\subsection{Training and Evaluation}

All models were trained on a single NVIDA Quadro RTX 4000 with one CPU. Running the Single-Task, Fine-tune, and Multi models takes 7 hours for each set of three iterations, and running the Multi\textsuperscript{\textit{Alt}} models takes 11 hours for each set of three iterations. Running the entire hyperparameter tuning setup with three iterations for each of the 16 primary experiment models and 32 data reduction models takes 7 days.

Most of the parameters in the models come from their pretrained language models. Models with BERT or MentalBERT bases have 110M parameters, and models with RoBERTa or MentalRoBERTa bases have 125M parameters.

To evaluate our models, we use the standard formula for binary F1, where the presence of either psychological stress or minority stress is the positive class. We also report the classification accuracy using the standard formula.

Our performance for each model on the development set is shown in Table~\ref{tab:dev-result}.

\subsection{Hyperparameter Tuning}

We replicate \citet{turcan-etal-2021-emotion}'s hyperparameter tuning process with the same parameter ranges. All models tune the learning rate and dropout probability, but Multi models also tune the lambda parameter for the multi-task loss function. The learning rate is tuned in the range $[10^{-6}, 10^{-3}]$. The dropout probability is tuned in the range $[0, 1]$, and $\lambda$ is tuned in the range $[0, 0.9]$ We tune the hyperparameters of each model using \texttt{ax}’s implementation of Bayesian optimization\footnote{https://github.com/facebook/Ax}, and we use the F1 score on the MStress development set as the criterion for optimization. We tune the models for a maximum of 20 epochs, with a patience of 5 epochs and a tolerance of 0.0001.

\subsection{Datasets}

\begin{table}[ht] 
\caption{The number of examples and label distribution for GoEmotions.}
\begin{tabular}{lll}
Label                         & \# of Examples & Proportion \\ \hline
\multicolumn{1}{l|}{Joy}      & 21,733         & 40.1\%                 \\
\multicolumn{1}{l|}{Neutral}  & 17,772         & 32.8\%                 \\
\multicolumn{1}{l|}{Anger}    & 7,022          & 12.9\%                 \\
\multicolumn{1}{l|}{Surprise} & 6,668          & 12.3\%                 \\
\multicolumn{1}{l|}{Sadness}  & 4,032          & 7.4\%                  \\
\multicolumn{1}{l|}{Disgust}  & 1,013          & 1.9\%                  \\
\multicolumn{1}{l|}{Fear}     & 929            & 1.7\%                  
\end{tabular}
\label{tab:ge-labels}
\end{table}

All our data consists of English posts and comments from Reddit.

The Dreaddit dataset \citep{turcan-mckeown-2019-dreaddit} consists of 3,553 segments of Reddit posts labeled with whether they contain expressions of psychological stress. 52.3\% of the data is labeled as expressing psychological stress, and 46.4\% is labeled as not expressing psychological stress. For our primary experiments, we use 2,122 examples for training. For the data reduction experiments, we use 1,591, 1,060, 530, and 212 examples for training the 75\%, 50\%, 25\%, and 10\% proportions respectively. For the data reduction experiments, we use 716 examples for development. For all evaluations of psychological stress detection, we use 715 examples for testing.

The MStress dataset \citep{saha_language_2019} consists of 350 Reddit posts from LGBTQ+ communities labeled with whether they contain expressions of minority stress. 41.4\% of the data is labeled as expressing minority stress, and 58.6\% is labeled as not expressing minority stress. For our experiments, we do not directly train on any of the minority stress data. We use 175 examples for development and 175 examples for testing.

The GoEmotions dataset \citep{demszky-etal-2020-goemotions} consists of 58,009 Reddit comments labeled with 28 emotion labels. We use a relabeling of the dataset provided by the authors that groups the emotions into 7 labels. Table~\ref{tab:ge-labels} provides the frequency of each of the 7 labels. For all experiments, we use 42,409 examples for training, 5,425 examples for development, and 5,426 examples for testing. 

We do not filter or remove any examples. The only preprocessing performed is the use of the use of \citet{wolf-etal-2020-transformers}'s tokenizers associated with each of the pretrained language models used.

\end{document}